\def\year{2018}\relax
\documentclass[letterpaper]{article} 
\usepackage{aaai18}  
\usepackage{times}  
\usepackage{helvet}  
\usepackage{courier}  
\usepackage{url}  
\usepackage{graphicx}  
\usepackage{epsfig}
\usepackage{amsmath}
\usepackage{amssymb}
\usepackage{multirow}
\usepackage{array}
\usepackage{booktabs}
\usepackage{amsthm}
\usepackage{subfigure}
\usepackage{float}

\usepackage{algorithm}
\usepackage{algorithmic}

\begin{document}
%
\title{Adversarial Discriminative Heterogeneous Face Recognition}
\author{Lingxiao Song, Man Zhang, Xiang Wu, Ran He$^*$\\
National Laboratory of Pattern Recognition, CASIA\\
Center for Research on Intelligent Perception and Computing, CASIA\\
Center for Excellence in Brain Science and Intelligence Technology, CAS\\
University of Chinese Academy of Sciences, Beijing 100190, China.\\
}
\maketitle


\begin{abstract}

The gap between sensing patterns of different face modalities remains a challenging problem in heterogeneous face recognition (HFR). This paper proposes an adversarial discriminative feature learning framework to close the sensing gap via adversarial learning on both raw-pixel space and compact feature space. This framework integrates cross-spectral face hallucination and discriminative feature learning into an end-to-end adversarial network. In the pixel space, we make use of generative adversarial networks to perform cross-spectral face hallucination. An elaborate two-path model is introduced to alleviate the lack of paired images, which gives consideration to both global structures and local textures. In the feature space, an adversarial loss and a high-order variance discrepancy loss are employed to measure the global and local discrepancy between two heterogeneous distributions respectively. These two losses enhance domain-invariant feature learning and modality independent noise removing.
Experimental results on three NIR-VIS databases show that our proposed approach outperforms state-of-the-art HFR methods, without requiring of complex network or large-scale training dataset.

\end{abstract}


\section{Introduction}

Face recognition research has been significantly promoted by deep learning techniques recently. But a persistent challenge remains to develop methods capable of matching heterogeneous faces that have large appearance discrepancy due to various sensing conditions. Typical heterogeneous face recognition (HFR) tasks conclude visual versus near infrared (VIS-NIR) face recognition~\cite{yi2007face_NIS-VIR,yi2009partial_NIS-VIR}, visual versus thermal infrared (VIS-TIR) face recognition~\cite{socolinsky2002TIR_analysis}, face photo versus face sketch~\cite{tang2004face_sketch,wang_tang2009photo-sketch}, face recognition across pose~\cite{huang2017beyond} and so on. VIS-NIR HFR is the most popular and representative task in HFR. This is because NIR imaging provides a low-cost and effective solution to acquire high-quality images under low-light scenarios. It is widely applied in surveillance systems nowadays. However, the popularization of NIR images is far from VIS images, and most face databases are enrolled in VIS domain. Consequently, the demand for face matching between NIR and VIS images grows gradually.

A major challenge of HFR comes from the gap between sensing patterns of different face modalities. In practice, human face appearance is often influenced by many factors, including identities, illuminations, viewing angles, expressions and so on. Among all the factors, identity difference accounts for intra-personal differences while the rest lead to inter-personal differences. A key effort for face recognition is to alleviate intra-personal differences while enlarge inter-personal differences. Specifically, in the heterogeneous case, the noise factors that cause inter-personal differences show diverse distributions in different modalities, e.g. various spectrum sensing distribution between VIS domain and NIR domain, leading to a more complex problem to preserving the identity relevance between different modalities.

\begin{figure}[t]
\begin{center}
\includegraphics[width=1.05\linewidth]{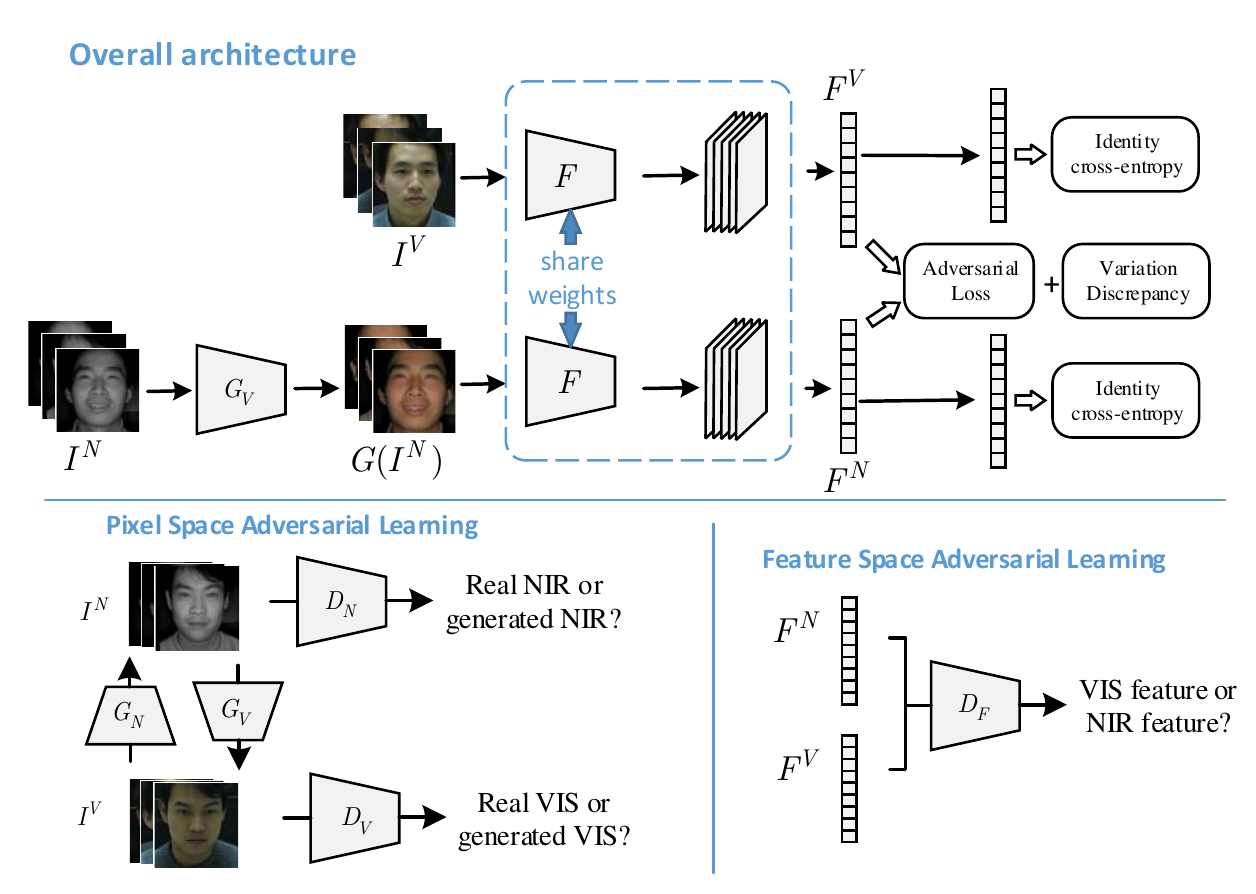}
\end{center}
   \caption{The proposed adversarial discriminative HFR framework. Adversarial learning is employed on both raw-pixel space and compact feature space.}
\label{fig:pipeline}
\end{figure}

A lot of research efforts have been devoted to eliminating the sensing gap~\cite{socolinsky2002TIR_analysis,yi2007face_NIS-VIR,li2013casia}. One straightforward approach to cope with the sensing gap is to transform heterogeneous data onto a common comparable space~\cite{lei2012coupled}. Another commonly used strategy is to map data from one modality to another~\cite{lei2008CCA_mapping,wang2009analysis,huang2013coupled}. Most of these methods only focus on minimizing the sensing gap, but not emphasize discrimination among different subjects, causing performance reduction when the number of subjects increases.

Another challenge for HFR is the lack of paired training data. General face recognition and hallucination have benefited a lot from the development of deep neural networks. However, the success of deep learning relies on large amount of labeled or paired training data to some extent. Although we can easily collect large-scale VIS images through the internet, it is hard to collect massive paired heterogeneous image data such as NIR images and TIR images. How to take the advantage of the powerful general face recognition to boost HFR and cross-spectral face hallucination is worth studying.

To address the above two issues, this paper proposes an adversarial discriminative feature learning framework for HFR by introducing adversarial learning on both raw-pixel space and compact feature space. Figure~\ref{fig:pipeline} is the pipeline of our approach. Cross-spectral face hallucination and discriminative feature learning are simultaneously considered in this  network. In the pixel space, we make use of generative adversarial networks (GAN) as a sub-network to perform cross-spectral face hallucination. An elaborate two-path model is introduced in this sub-network to alleviate the lack of paired images, which gives consideration to both global structures and local textures and results in a better visual result. In the feature space, an adversarial loss and a high-order variance discrepancy loss are employed to measure the global and local discrepancy between two heterogeneous feature distributions respectively. These two losses enhance domain-invariant feature learning and modality independent noise removing. Moreover, we implement all these global and local information in an end-to-end adversarial network, resulting in relatively compact 256 dimensional features. Experimental results show that our proposed adversarial approach not only outperforms state-of-the-art HFR methods but also can generate photo-realistic VIS images from NIR images, without requiring of complex network or large-scale training dataset. The results also suggest that the joint hallucination and feature learning is helpful to reduce the sensing gap.

The main contributions are summarized as follows,
\begin{itemize}
\item A cross-spectral face hallucination framework is embedded as a sub-network in adversarial learning based on GAN. A two-path architecture is presented to cope with the absence of well aligned image pairs and improve face image quality.
\item An adversarial discriminative feature learning strategy is presented to seek domain-invariant features. It aims at eliminating the heterogeneities in compact feature space and reducing the discrepancy between different modalities in terms of both local and global distributions.
\item Extensive experimental evaluations on three challenging HFR databases demonstrate the superiority of the proposed adversarial method, especially taking feature dimension and visual quality into consideration.
\end{itemize}

\section{Related Work}


What makes heterogeneous face recognition different from general face recognition is that we need to place data from different domains to the same space, only by which the measurement between heterogeneous data can make sense.


A kind of approaches uses data synthesis to map data from one modality into another. Thus the similarity relationship of heterogeneous data from different domain can be measured. In \cite{liu2005nonlinear}, a local geometry preserving based nonlinear method is proposed to generate pseudo-sketch from face photo. In~\cite{lei2008CCA_mapping}, they propose a canonical correlation analysis (CCA) based multi-variate mapping algorithm to reconstruct 3D model from a single 2D NIR image. In~\cite{wang_tang2009photo-sketch}, multi-scale Markov Random Fields (MRF) models are extend to synthesize sketch drawing from given face photo and vice versa. In \cite{wang2009analysis}, a cross-spectrum face mapping method is proposed to transform NIR and VIS data to another type. Many works~\cite{wang2012sketch_synthesis,juefei2015cvprw_nir} resort to coupled or joint dictionary learning to reconstruct face images and then perform face recognition.
However, large amount of pairwise multi-view data are essential for these methods based on data synthesis, making it very difficult to collect training images. In~\cite{lezama2016not}, they design a patch mining strategy to collect aligned image patches, and then produce VIS faces from NIR images through a deep learning approach.

Another kind of methods deals with heterogeneous data by projecting them to a common latent space respectively, or learn modality-invariant features that are robust to domain transfer. 
In~\cite{lin_tang2006}, 
Common Discriminant Feature Extraction (CDFE) is proposed to transform data to a common feature space, which takes both inter-modality discriminant information and intra-modality local consistency into consideration.
\cite{liao2009heterogeneous} use DoG filtering as preprocessing for illumination normalization, and then employ Multi-block LBP (MB-LBP) to encode NIR as well as VIS images.
\cite{klare2010heterogeneous} further combine HoG features to LBP descriptors, and utilize sparse representation to improve recognition accuracy.
\cite{goswami2011evaluation} incorporate a series of preprocessing methods to do normalization, then combine Local Binary Pattern Histogram (LBPH) representation with LDA to extract robust features.
In~\cite{zhang2011coupled}, a coupled information-theoretic projection method is proposed to reduce the modality gap by maximizing the mutual information between photos and sketches in the quantized feature spaces.
In~\cite{lei2012coupled}, a coupled discriminant analysis method is suggested that involves the locality information in kernel space.
In~\cite{huang2013regularized}, a regularized discriminative spectral regression (DSR) method is developed to map heterogeneous data into the same latent space.
In~\cite{hou2014domain}, a domain adaptive self-taught learning approach is developed to derive a common subspace.
In ~\cite{zhu2014matching}, Log-DoG filtering is involved with local encoding and uniform feature normalization to reduce heterogeneities between VIS and NIR images.
~\cite{shao2017cross} propose a hierarchical hyperlingual-words (Hwords) to capture high-level semantics across different modalities, and a distance metric through the hierarchical structure of Hwords is presented accordingly.

Recently, many works attempt to address the cross-modal matching problem by deep learning methods benefitting from the development of deep learning.
In~\cite{yidong2015shared} , Restricted Boltzmann Machines (RBMs) is used to learn a shared representation between different modalities.
In~\cite{liu2016transferring}, the triplet loss is applied to reduce intra-class variations among different modalities as well as augment the number of training sample pairs.
\cite{kan2016cross-view} develop a multi-view deep network that is made up of view-specific sub-network and common sub-network, in which the view-specific sub-network attempts to remove view-specific variations while the common sub-network seeks for common representation shared by all views.
In~\cite{he2017idr}, subspace learning and invariant feature extraction are combined into CNNs. This method obtains the state-of-the-art HFR result on CASIA NIR-VIS 2.0 database.
%



As mentioned before, our work is also related to the famous adversarial learning. GAN~\cite{goodfellow2014generative} has achieved great success in many computer vision applications including image style transfer~\cite{zhu2017unpaired,pix2pix2016}, image generation~\cite{shrivastava2016learning,huang2017beyond} , image super-resolution~\cite{ledig2016photo}, object detection~\cite{li2017perceptual,wang2017fast}.
Adversarial learning provides a simple yet efficient way to fit target distribution via the min-max two-player game between generator and discriminator. Motivated by this, we introduce adversarial learning in NIR-VIS face hallucination and domain-invariant feature learning, aiming at closing the sensing gap of heterogeneous data in pixel space and feature space simultaneously.


\section{The Proposed Approach}
In this section, we present a novel framework for the cross-modal face matching problem based on adversarial discriminative learning. We first introduce the overall architecture,  and then describe the cross-spectral face hallucination and the adversarial discriminative feature learning separately.

\subsection{Overall Architecture}

The goal of this paper is to design a framework that enables learning of domain-invariant feature representations for images from different modalities, i.e. VIS face images $I^V$ and NIR face images $I^N$.

We can easily get numerous VIS face images for training thanks to the prosperous of social network. In most circumstances, face recognition approaches are trained with VIS face images, which cannot achieve full performance when handling with NIR images. Besides, it is necessary to archive all processed images for most face recognition systems in real-world applications. However, NIR face images are much harder to distinguish by humans comparing with VIS faces. A feasible way is to convert NIR face images into VIS spectrum. Thus, we employ a GAN to perform cross-spectral face hallucination, aiming at better fitting the VIS-based face models as well as producing VIS-like images that are friendly to human eyes.

However, we find that it is insufficient that only transferring NIR images into VIS spectrum in NIR-VIS HFR. A reasonable explanation is that NIR images are distinct with VIS images not just on imaging spectrum. For example, NIR face images often have darker or blurrier outlines due to the distance limit of the near-infrared illumination. The special way of imaging for NIR images makes the noise factors that cause inter-personal differences show diverse distributions compared to the VIS images. Hence, an adversarial discriminative feature learning strategy is proposed in our approach to reduce heterogeneities between VIS and NIR images.

To summarize, the proposed approach consists of two key components (shown in Fig.~\ref{fig:pipeline}): cross-spectral face hallucination and adversarial discriminative feature learning. These two components try to eliminate the gap between different modalities in raw-pixel space and compact feature space respectively.

\subsection{Cross-spectral Face Hallucination}
The outstanding performance of GAN in fitting data distribution has significantly promoted many computer vision applications such as image style transfer~\cite{zhu2017unpaired,pix2pix2016}.
Motivated by its remarkable success, we employ GAN to perform the cross-spectral face hallucination that converting NIR face images into VIS spectrum.

A major challenge in NIR-VIS image converting is that image pairs are not aligned accurately in most databases. Even though we can align images based on facial landmarks, the pose and facial expression of the same subject still vary quite a lot. Therefore, we build our cross-spectral face hallucination models based on the CycleGAN framework~\cite{zhu2017unpaired}, which can handle unpaired image translation tasks.
As illustrated in Fig.~\ref{fig:pipeline}, a pair of generators ${G_V}:{I^N} \to {I^V}$ and ${G_N}:{I^V} \to {I^N}$ are introduced to achieve opposite transformation, with which we can construct mapping cycles between VIS and NIR domain. Associated with these two generators, $D_V$ and $D_N$ aim to distinguish between real images $I$ and generated images $G(I)$ correspondingly.

Generators and discriminators are trained alternatively toward adversarial goals, following the pioneering work of~\cite{goodfellow2014generative}. The adversarial losses for generator and discriminator are shown in Eq.~\ref{L_adv_G} and Eq.~\ref{L_adv_D} respectively.
\begin{equation}\label{L_adv_G}
\begin{split}
{L_{G-adv}} =  - {{\mathbb{E}}_{{I} \sim P\left( {{I}} \right)}}\log {D}\left( {G\left( {{I}} \right)} \right),
\end{split}
\end{equation}
\begin{equation}\label{L_adv_D}
\begin{split}
{L_{D-adv}} &={{\mathbb{E}}_{{I^{'}} \sim P\left( {{I^{'}}} \right)}}\log {D}( {1 - {I^{'}}} ) \\
&+{{\mathbb{E}}_{{I} \sim P\left( {{I}} \right)}}\log {D}\left( {G\left( {I} \right)} \right),
\end{split}
\end{equation}
where $I$ and $I^{'}$ are images from different modalities.

In the CycleGAN framework, an extra cycle consistency loss $L_{cyc}$ is introduced to guarantee consistency between input images and the reconstructed images, e.g. $I^N$ vs. ${G_{N}}({G_{V}}(I^N))$ and ${G_{V}}({G_{N}}(I^V))$. $L_{cyc}$ is calculated as
\begin{equation}\label{L_cyc}
\begin{split}
{L_{cyc}} = {E_{I \sim P\left( I \right)}}{\left\| {I - F\left( {G\left( I \right)} \right)} \right\|_1},
\end{split}
\end{equation}
where $F$ is the opposite generator to $G$. In our cross-spectral face hallucination case, if $G$ is used to transfer VIS faces into NIR spectrum, then $F$ is used to transfer NIR faces into VIS spectrum.

\begin{figure}[t]
\begin{center}
\includegraphics[width=1.0\linewidth]{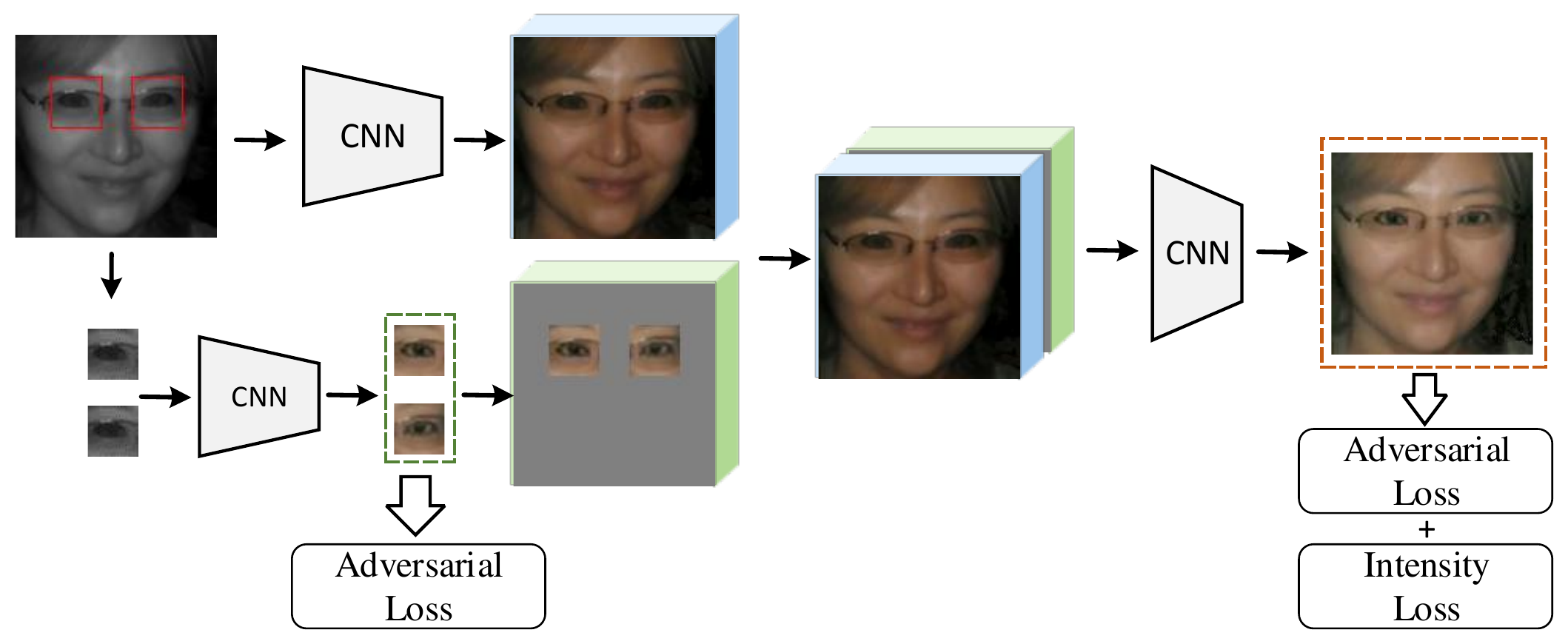}
\end{center}
   \caption{The proposed two-path architecture used in cross-spectral face hallucination.}
\label{fig:Two_path_model}
\end{figure}

We find that a single generator is hard to synthesize high quality cross-spectral images with both global structures and local details are well reconstructed. A possible explanation is that convolutional filters are shared across all the spatial locations, which are seldom suitable for recovering global and local information at the same time. Therefore, we employ a two-path architecture as shown is Fig.~\ref{fig:Two_path_model}. Since the periocular regions show special correspondences between NIR images and VIS images diverse from other facial areas, we add a local path around eyes so as to precisely recover details of the periocular regions.

Because VIS images and NIR images mainly have difference in light spectrum, the structure information should be preserved after cross-spectral translations. Similar to~\cite{lezama2016not}, we choose to represent the input and output images in YCbCr space, for which the luminance component Y encode most structure information as well as identity information. An luminance-preserving term is adopted in the global path to enforce structure consistency:
\begin{equation}\label{L_intensity}
\begin{split}
L_{intensity} = {{\mathbb{E}}_{I \sim P\left( I \right)}}{\left\| {Y\left( I \right) - Y\left( {G\left( I \right)} \right)} \right\|_1}
\end{split}
\end{equation}
in which $Y(.)$ stands for the Y channel of images in YCbCr space.

To sum up, the full objective for generators $G_V, G_N$ is:
\begin{equation}\label{L_G_full}
\begin{split}
{L_G} = {L_{G - adv}} + {\alpha _1}{L_{cyc}} + {\alpha _2}{L_{intensity}}
\end{split}
\end{equation}
where $\alpha_1$ and $\alpha_2$ are loss weight coefficients.

\subsection{Adversarial Discriminative Feature Learning}

In this section, we propose a simple way to learn domain-invariant face representations using adversarial discriminative feature learning strategy. Ideal face feature extractor should be capable of alleviating the discrepancy caused by different modalities, while keeping discriminant among different subjects.

\subsubsection{Adversarial Loss}
As mentioned above, GAN has strong ability of fitting target distribution via the simple min-max two-player game. In this section, we use GAN in cross-view feature learning so as to eliminate domain discrepancy in feature-level. As demonstrated in Fig.~\ref{fig:pipeline}, an extra discriminator $D_F$ is employed to act as the adversary to our feature extractor. $D_F$ outputs a scalar value that indicates the probability of belonging to VIS feature space. The adversarial loss of our feature extractor takes the form:

\begin{equation}\label{L_adv_DF}
\begin{split}
{L_{F-{adv}}} = - {{\text{E}}_{{I^N} \sim P\left( {{I^N}} \right)}}\log {D_f }\left( {{F}\left( {{G_V }\left( {{I^N}} \right)} \right)} \right)
\end{split}
\end{equation}

By enforcing the fitting of NIR feature distribution to VIS feature distribution, we can remove the noise factors accounting for domain discrepancy.
Since the adversarial loss is used to eliminate the discrepancy between distributions of heterogeneous data in a global view without taking local discrepancy into consideration, and distributions in each modalities consist of many sub-distributions of different subjects, the local consistency may not be well preserved.

\subsubsection{Variance Discrepancy}

Similar to the conventional domain adaptation tasks~\cite{long2016RTN,zellinger2017CMD}, we want to bridge two different domains by learning domain-invariant feature representations in HFR. But HFR faces more challenges. First, HFR needs to match the same subject or instance rather than the same class, and distinguishe two different subjects that belong to the same class in most domain adaptation tasks. Second, there is no upper limit of the number of subject classes, the majority of which are not appeared in training phase. Fortunately, unlike these unsupervised domain adaptation tasks, label information in the target domain is supported in HFR, which can supervise the discriminative feature learning.

The usage of adversarial loss can only handle partial intra-personal difference caused by modality transfer, but not the modality-independent noise factors.
Considering that the feature distribution of the same subject should be as close as possible ideally, we employ the class-wise variance discrepancy (CVD) to enforce the consistency of subject-related variation with the guide of identity label information:

\begin{equation}\label{moment}
\begin{split}
\sigma \left( F \right) = {\mathbb{E}}\left( {{{\left( {F - {\mathbb{E}}\left( F \right)} \right)}^2}} \right),
\end{split}
\end{equation}
\begin{equation}\label{L_moment}
\begin{split}
{L_{\text{CVD}}} = \sum\limits_{c = 1}^C {{\mathbb{E}}\left( {{{\left\| {\sigma \left( {{F_c}^V} \right) - \sigma \left( {{F_c}^N} \right)} \right\|}_2}} \right)}
\end{split}
\end{equation}
where $\sigma(.)$ is the variance function, and the ${F_c}^V$, ${F_c}^N$ denote feature observations belonging to the $c-$th class in VIS and NIR domain respectively.

\subsubsection{Cross-Entropy Loss}
 As the adversarial loss and the variance discrepancy penalties cannot ensure the inter-class diversity which exists in both the source domain and the target domain, we further employ the common-used classification architecture to enforce the discrimination and compactness of the learned feature. Empirical error of all samples is minimized as
\begin{equation}\label{L_softmax}
\begin{split}
{L_{cls}} = \frac{1}{{\left| N \right| + \left| V \right|}}\sum\limits_{i \in \{ N,V\} } {{\mathcal{L}}\left( {W{F_i},{y_i}} \right)}
\end{split}
\end{equation}
where $W$ is the parameter for softmax normalization, and $\mathcal{L}\left( { \cdot , \cdot } \right)$ is the cross-entropy loss function.

The final loss function is a weighted sum of all the losses defined above: $L_{adv}$ to remove the modality gap, $L_{\text{CVD}}$ to guarantee intra-class consistency, and $L_{cls}$ to preserve identity discrimination.
\begin{equation}\label{L_final}
\begin{split}
L = {L_{F-{adv}}} + {\lambda _1}{L_{\text{CVD}}} + {\lambda _2}{L_{cls}}
\end{split}
\end{equation}

\section{Experiments}
In this section, we evaluate the proposed approach on three NIR-VIS databases. The databases and testing protocols are introduced firstly. Then, the implementation details is presented. Finally, comprehensive experimental analysis is conducted among the comparison with related works.
\subsection{Datasets and Protocols}

\textbf{The CASIA NIR-VIS 2.0 face database~\cite{li2013casia}}.
It is so far the largest as well as the most challenging public face database across NIR and VIS spectrum. Its challenge contains large variations of the same identity, expression, pose and distance. The database collects 725 subjects, each with 1-22 VIS and 5-50 NIR images. All images in this database are randomly gathered, and no one-to-one correspondence between NIR and VIS images. In our experiments, we follow the View 2 of the standard protocol defined in~\cite{li2013casia}, which is used for performance evaluation. There are 10-fold experiments in View 2, where each fold contains non-overlapped training and testing lists. There are about 6,100 NIR images and 2,500 VIS images from about 360 identities for training in each fold. In the testing phase, cross-view face verification is taken between the gallery set of 358 VIS image belonging to different subjects, and the probe set of over 6,000 NIR images from the same 358 identity. The Rank-1 identification rate and the ROC curve are used as evaluation criteria.

\textbf{The BUAA-VisNir face database~\cite{Huang2012Buaa}}.
This dataset is made up of 150 subjects with 40 images per subject, among which there are 13 VIS-NIR pairs and 14 VIS images in different illumination. Each VIS-NIR image pairs are captured synchronously using a single multi-spectral camera. The paired images in the BUAA-VisNir dataset vary in poses and expressions. Following the testing protocol proposed in~\cite{shao2017cross}, 900 images of 50 subjects are randomly selected for training, and the other 100 subjects make up the testing set. It is worth noted that the gallery set contains only one VIS image of each subject. Therefore, a testing set of 100 VIS images and 900 NIR images are organized. We report the Rank-1 accuracy and the ROC curve according to the protocol.

\textbf{The Oulu-CASIA NIR-VIS facial expression database~\cite{chen2009learning}}.
Videos of 80 subjects with six typical expressions and three different illumination conditions are captured in both NIR and VIS imaging systems in this database. We take cross-spectral face recognition experiments following the protocols in~\cite{shao2017cross}, where only images from the normal indoor illumination are used. In each expression, eight face images are randomly selected such that 48 VIS images and 48 NIR images of each subject are used. Based on the protocol in~\cite{shao2017cross}, the training set and testing set contain 20 subjects respectively, resulting in a total of 960 gallery VIS images and 960 NIR probe images in testing phase. Similar to the above two datasets, the Rank-1 accuracy and the ROC curve are reported.

%

\begin{table*}[htbp]
  \centering
  \caption{Experimental results for the 10-fold face verification tasks on the CASIA NIR-VIS 2.0 database of the proposed method.}
    \begin{tabular}{l|c|c|c|c}
    \toprule
          & Rank-1 acc.(\%) & VR@FAR=1\%(\%) & VR@FAR=0.1\%(\%) & VR@FAR=0.01\%(\%) \\
    \midrule
    \midrule
    Basic model & $87.16\pm0.45$ & $89.65\pm0.89$ & $72.06\pm1.38$ & $48.25\pm2.68$  \\
    Softmax     & $95.89\pm0.75$ & $98.26\pm0.48$ & $93.25\pm1.14$ & $75.13\pm3.02$  \\
    \midrule
    ADFL w/o $L_{adv}$  & $96.56\pm0.63$ & $98.56\pm0.27$ & $95.24\pm0.36$ & $81.69\pm1.77$  \\
    ADFL w/o $L_{\text{CVD}}$ & $ 97.34\pm0.53$ & $98.95\pm0.14$ & $96.88\pm0.40$ &  $85.83\pm3.02$ \\
    \midrule
    Hallucination  & $90.56\pm0.86$ & $92.95\pm0.20$ & $81.17\pm0.42$ & $62.24\pm2.77$  \\
    ADFL & $97.81\pm0.29$ & $99.04\pm0.21$ & $\textbf{97.21}\pm0.34$ &  $\textbf{88.11}\pm3.09$ \\
    Hallucination + ADFL & $\textbf{98.15}\pm0.34$ & $\textbf{99.12}\pm0.15$ & $97.18\pm0.48$ &  $87.79\pm2.33$ \\
    \bottomrule
    \end{tabular}%
  \label{tab:res_ours}%
\end{table*}%
\subsection{Implementation Details}

%
%

\begin{figure}[t]
\begin{center}
\includegraphics[width=1.0\linewidth]{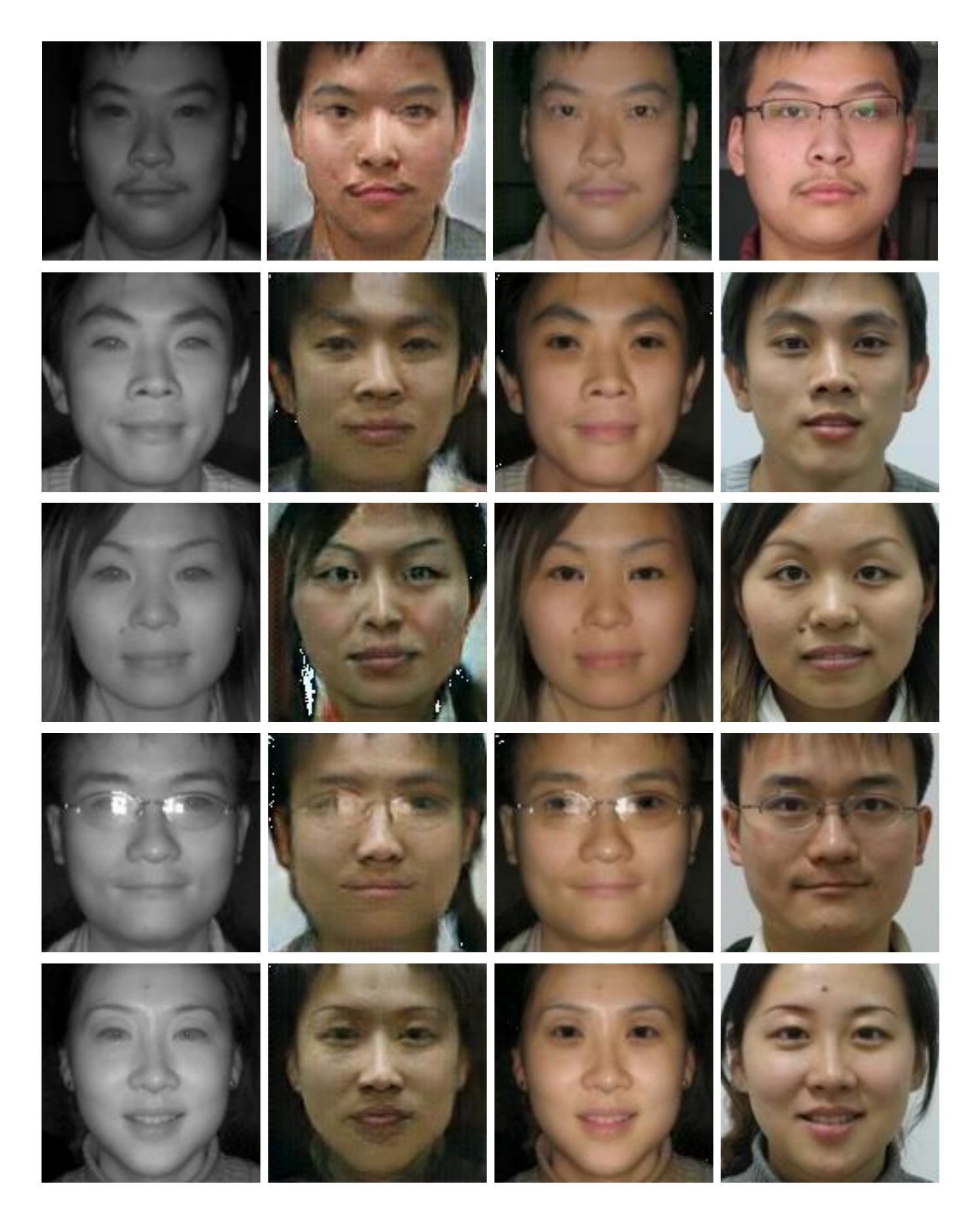}
\end{center}
   \caption{Results of the cross-spectral face hallucination. From left to right, the input NIR images, generated VIS images by cycleGAN, generated VIS images by the proposed cross-spectral face hallucination framework, and corresponding VIS images of the same subjects.}
\label{fig:face_generate_res}
\end{figure}

\textbf{Training data}.
Our cross-spectral hallucination network is trained on the CASIA NIR-VIS 2.0 face dataset. Note that the label annotation is not involved in the training of face hallucination module, therefore it would not affect the reliability of our following HFR tests. The feature extraction network is pre-trained on the MS-Celeb-1M dataset~\cite{guo2016ms}, and finetuned on each testing datasets respectively. All the face images are normalized by similarity transformation using the locations of two eyes, and then cropped to $144 \times 144$ size, of which $128 \times 128$ sized sub images are selected by random cropping in training and center cropping in testing. For the local-path, $32 \times 32$ patches are cropped around two eyes, and then flipped to the same side. As mentioned above, in the cross-spectral hallucination module, images are encoded in YCbCr space. In the feature extraction step, grayscale images are used as input.

\textbf{Network architecture}.
Our cross-spectral hallucination networks take the architecture of ResNet~\cite{he2016resnet}, where the global-path is comprised of 6 residual blocks and the local-path contains 3 residual blocks. Output of the local-path is feed to the global-path before the last block. In the adversarial discriminative feature learning module, we employ the model-B of the Light CNN~\cite{wu2015lightened} as our basic model, which includes 9 convolution layers, 4 max-pooling and one fully-connected layer. Parameters of the convolution layers are shared across the VIS and NIR channels as shown in Fig.~\ref{fig:pipeline}. The output feature dimension of our approach is 256, which is relatively compact comparing with other state-of-the-art face recognition networks.

\subsection{Experimental Results}

\subsubsection{Face Hallucination Results}

Fig.~\ref{fig:face_generate_res} shows some examples generated by our cross-spectral hallucination framework. We report the results of cycleGAN~\cite{zhu2017unpaired} for comparison. As shown in Fig.~\ref{fig:face_generate_res}, the results of cycleGAN are not satisfying, which may caused by the lack of strong constraint such as the proposed $L_{intensity}$. Note that our method can accurately recover details of the VIS faces, e.p. eyeballs, mouths and hairs. Specifically, the periocular regions are well transformed to VIS-like faces in which eyeballs are distinguishable.
Results in Fig.~\ref{fig:face_generate_res} demonstrate the ability of our cross-spectral hallucination framework to generate photo-realistic VIS images from NIR inputs, with both global structure and local details are well preserved.



\subsubsection{Results on the CASIA NIR-VIS 2.0 database}


Table~\ref{tab:res_ours} shows results of the proposed approach with different settings. We report mean value and standard deviation of Rank-1 identification rate, verification rates at $1\%$, $0.1\%$, $0.01\%$ false accept rate (VR@FAR=$1\%$, VR@FAR=$0.1\%$, VR@FAR=$0.01\%$) for a detailed analysis.
We evaluate the performance obtained by our method in different settings, including cross-spectral hallucination, ADFL and hallucination $+$ ADFL. In order to validate the effectiveness of $L_{adv}$ and $L_{\text{CVD}}$, we report results of removing one of them respectively.
The cross-spectral hallucination brings a performance gain for  about $3\%$ in Rank-1 accuracy as well as VR@FAR=$1\%$, addressing that the cross-spectral image transfer helps to close the sensing gap between different modalities.
Obviously, significant improvements can be observed when the proposed ADFL is used.
Since supervision signals are introduced in the ADFL, it has stronger capacity than cross-spectral hallucination to boost the HFR accuracy.
Both the adversarial loss and the variance discrepancy help to improve the recognition performance according to results of w/o $L_{adv}$ and w/o $L_{\text{CVD}}$.
When the cross-spectral hallucination and the adversarial discriminative learning strategies are applied together,  the best performance is obtained.

\begin{table}
  \centering
  \caption{Experimental results for the 10-fold face verification tasks on the CASIA NIR-VIS 2.0 database.}
    \begin{tabular}{l|c|c|c}
    \toprule
          & Rank-1 & FAR=0.1\% & Dim. \\
    \midrule
    \midrule
    PCA+Sym+HCA(2013) & $23.70$ & 19.27 & -\\
    LCFS(2015) & $35.40$ & 16.74 & -\\
    CDFD(2015) & $65.8$ & 46.3  & -\\
    CDFL(2015) & $71.5$ & 55.1  & 1000 \\
    Gabor+RBM(2015) & $86.16$ & $81.29$ & 14080 \\
    Recon.+UDP(2015) & $78.46$ & 85.80  & - \\
    $\text{H}^2(\text{LBP}_3)$(2016) & 43.8 &10.1 & - \\
    COTS+Low-rank(2017) & 89.59 & - & 1024 \\
    IDR(2017) & $97.33$ & $95.73$  & \textbf{128} \\
    \midrule
    Ours   & \textbf{98.15} & $\textbf{97.18}$ & 256\\
    \bottomrule
    \end{tabular}%
  \label{tab:res_nir2.0}%
\end{table}%

We also compare the proposed approach with both conventional and state-of-the-art deep learning based NIR-VIS face recognition methods: PCA+Sym+HCA~\cite{li2013casia}, learning coupled feature space (LCFS)~\cite{jin2015coupled}, coupled discriminant face descriptor(CDFD)~\cite{jin2015coupled,wangkaiye2013iccv},coupled discriminant feature learning (CDFL)~\cite{jin2015coupled}, Gabor+RBM~\cite{yidong2015shared}, NIR-VIS reconstruction+UDP~\cite{juefei2015cvprw_nir}, COTS+Low-rank~cite{lezama2016not} and Invariant Deep Representation (IDR)~\cite{he2017idr}.
%
The experimental results are consolidated in Table~\ref{tab:res_nir2.0}.
We can see that deep learning based HFR methods perform much better than conventional approaches.
The proposed method improves the previous best Rank-1 accuracy and VR@FAR=$0.1\%$, which are obtained by IDR in ~\cite{he2017idr}, from $97.33\%$ to $98.14\%$ and $95.73\%$ to $97.18\%$ respectively. All of these results suggest that our method is effective for the NIR-VIS recognition problem.

\subsubsection{Results on the BUAA-VisNir face database}

\begin{table}
  \centering
  \caption{Experimental results on the BUAA-VisNir Database.}
    \begin{tabular}{l|c|c|c}
    \toprule
          & Rank-1 & FAR=1\% & FAR=0.1\% \\
    \midrule
    \midrule
    MPL3(2009) &  53.2 & 58.1 & 33.3\\
    KCSR(2009) & 81.4 & 83.8 & 66.7\\
    KPS(2013) & 66.6 & 60.2 & 41.7\\
    KDSR(2013) &  83.0 & 86.8 & 69.5\\
    $\text{H}^2(\text{LBP}_3)$(2017) & 88.8 & 88.8 &73.4 \\
    IDR(2017) & 94.3 & 93.4 & 84.7 \\
    \midrule
    Basic model & 92.0&  91.5& 78.9 \\
    Softmax & 94.2 & 93.1 & 80.6 \\
    ADFL w/o $L_{\text{CVD}}$ & 94.8& 92.2& 83.9 \\
    ADFL w/o $L_{adv}$ & 94.9& 94.5& 87.7 \\
    ADFL  & \textbf{95.2}& \textbf{95.3} & \textbf{88.0} \\
    \bottomrule
    \end{tabular}%
  \label{tab:res_buaa}%
\end{table}%
We compare the proposed approach with MPL3~\cite{chen2009learning}, KCSR~\cite{lei2009coupled}, KPS~\cite{lei2009coupled}, KDSR~\cite{huang2013regularized} and $\text{H}^2(\text{LBP}_3$~\cite{shao2017cross}. The results of these comparing methods are from~\cite{shao2017cross}.
Table~\ref{tab:res_buaa} shows the Rank-1 accuracy and verification rate of each method.
Profit from the powerful large-scale training data, the basic model achieves really good performance that is better than most of the comparing methods. We can see that performance can be further improved when adversarial loss and variance discrepancy are introduced.
 Particularly, without the constraint of variance consistency, the verification rate drops dramatically at low FAR. This phenomenon demonstrates the effectiveness of variance discrepancy in removing intra-subject variations. Finally, the proposed ADFL acquires the best performance.

\subsubsection{Results on the Oulu-CASIA NIR-VIS facial expression database}
Results on the Oulu-CASIA NIR-VIS are presented in Table\ref{tab:res_oulu}, in which the results of these comparing methods are from~\cite{shao2017cross}. Similar to results on the BUAA-VisNir database, our proposed ADFL further boosts the performance beyond the powerful basic model. We observe that the adversarial loss contributes little to this database since the training set of Oulu-CASIA NIR-VIS database only contains 20 subjects and is relatively small-scale. So it is easy for the powerful Light CNN to learn good feature extractor for such a small dataset with the guidance of softmax loss.
Besides, the variance discrepancy still shows great capability in promoting verification rate at low FAR. These results demonstrate the superiority of our method.

\begin{table}
  \centering
  \caption{Experimental results on Oulu-CASIA NIR-VIS Database.}
    \begin{tabular}{l|c|c|c}
    \toprule
          & Rank-1 & FAR=1\% & FAR=0.1\% \\
    \midrule
    \midrule
    MPL3(2009) &  48.9 & 41.9 & 11.4\\
    KCSR(2009) & 66.0 & 49.7 & 26.1\\
    KPS(2013) & 62.2 & 48.3 & 22.2\\
    KDSR(2013) &  66.9 & 56.1 &31.9\\
    $\text{H}^2(\text{LBP}_3)$(2017) & 70.8 &62.0 &33.6 \\
    IDR(2017) & 94.3 & 73.4 & 46.2 \\
    \midrule
    Basic model & 92.2&  80.3& 53.1 \\
    Softmax & 93.0 & 80.9 & 56.1 \\
    ADFL w/o $L_{\text{CVD}}$ & 93.1 &81.2 &55.0 \\
    ADFL w/o $L_{adv}$ &92.7 & \textbf{83.5} & 60.6  \\
    ADFL & \textbf{95.5}& 83.0 & \textbf{60.7}  \\
    \bottomrule
    \end{tabular}%
  \label{tab:res_oulu}%
\end{table}%

%
%
%
%
%
%
%
%
%
%

\section{Conclusions}
In this paper, we focus on the VIS-NIR face verification problem. An adversarial discriminative feature learning framework is developed  by introducing adversarial learning in both raw-pixel space and compact feature space. In the raw-pixel space, the powerful generative adversarial network is employed to perform cross-spectral face hallucination, using a two-path architecture that is carefully designed to alleviate the absence of paired images in NIR-VIS transfer. As for the feature space, we utilize the adversarial loss and a high-order variance discrepancy loss to measure the global and local discrepancy between feature distributions of heterogeneous data respectively. The proposed cross-spectral face hallucination and adversarial discriminative learning are embedded in an end-to-end adversarial network, resulting in a compact 256-dimensional feature representation. Experimental results on three challenging NIR-VIS face databases demonstrate the effectiveness of the proposed method in NIR-VIS face verification.

\bibliography{HFR}

\begin{thebibliography}{}

\bibitem[\protect\citeauthoryear{Chen \bgroup et al\mbox.\egroup
  }{2009}]{chen2009learning}
Chen, J.; Yi, D.; Yang, J.; Zhao, G.; Li, S.~Z.; and Pietikainen, M.
\newblock 2009.
\newblock Learning mappings for face synthesis from near infrared to visual
  light images.
\newblock In {\em CVPR},  156--163.

\bibitem[\protect\citeauthoryear{Goodfellow \bgroup et al\mbox.\egroup
  }{2014}]{goodfellow2014generative}
Goodfellow, I.; Pouget-Abadie, J.; Mirza, M.; Xu, B.; Warde-Farley, D.; Ozair,
  S.; Courville, A.; and Bengio, Y.
\newblock 2014.
\newblock Generative adversarial nets.
\newblock In {\em NIPS},  2672--2680.

\bibitem[\protect\citeauthoryear{Goswami \bgroup et al\mbox.\egroup
  }{2011}]{goswami2011evaluation}
Goswami, D.; Chan, C.~H.; Windridge, D.; and Kittler, J.
\newblock 2011.
\newblock Evaluation of face recognition system in heterogeneous environments
  (visible vs nir).
\newblock In {\em ICCVW},  2160--2167.

\bibitem[\protect\citeauthoryear{Guo \bgroup et al\mbox.\egroup
  }{2016}]{guo2016ms}
Guo, Y.; Zhang, L.; Hu, Y.; He, X.; and Gao, J.
\newblock 2016.
\newblock Ms-celeb-1m: A dataset and benchmark for large-scale face
  recognition.
\newblock In {\em ECCV},  87--102.

\bibitem[\protect\citeauthoryear{He \bgroup et al\mbox.\egroup
  }{2016}]{he2016resnet}
He, K.; Zhang, X.; Ren, S.; and Sun, J.
\newblock 2016.
\newblock Deep residual learning for image recognition.
\newblock In {\em CVPR},  770--778.

\bibitem[\protect\citeauthoryear{He \bgroup et al\mbox.\egroup
  }{2017}]{he2017idr}
He, R.; Wu, X.; Sun, Z.; and Tan, T.
\newblock 2017.
\newblock Learning invariant deep representation for nir-vis face recognition.
\newblock In {\em AAAI},  2000--2006.

\bibitem[\protect\citeauthoryear{Hou, Yang, and Wang}{2014}]{hou2014domain}
Hou, C.-A.; Yang, M.-C.; and Wang, Y.-C.~F.
\newblock 2014.
\newblock Domain adaptive self-taught learning for heterogeneous face
  recognition.
\newblock In {\em ICPR},  3068--3073.

\bibitem[\protect\citeauthoryear{Huang and Frank~Wang}{2013}]{huang2013coupled}
Huang, D.-A., and Frank~Wang, Y.-C.
\newblock 2013.
\newblock Coupled dictionary and feature space learning with applications to
  cross-domain image synthesis and recognition.
\newblock In {\em ICCV},  2496--2503.

\bibitem[\protect\citeauthoryear{Huang \bgroup et al\mbox.\egroup
  }{2013}]{huang2013regularized}
Huang, X.; Lei, Z.; Fan, M.; Wang, X.; and Li, S.~Z.
\newblock 2013.
\newblock Regularized discriminative spectral regression method for
  heterogeneous face matching.
\newblock {\em ITIP} 22(1):353--362.

\bibitem[\protect\citeauthoryear{Huang \bgroup et al\mbox.\egroup
  }{2017}]{huang2017beyond}
Huang, R.; Zhang, S.; Li, T.; and He, R.
\newblock 2017.
\newblock Beyond face rotation: Global and local perception gan for
  photorealistic and identity preserving frontal view synthesis.
\newblock In {\em ICCV}.

\bibitem[\protect\citeauthoryear{Huang, Sun, and Wang}{2012}]{Huang2012Buaa}
Huang, D.; Sun, J.; and Wang, Y.
\newblock 2012.
\newblock The {BUAA}-{V}is{N}ir face database instructions.
\newblock Technical Report IRIP-TR-12-FR-001, Beihang University, Beijing,
  China.

\bibitem[\protect\citeauthoryear{Isola \bgroup et al\mbox.\egroup
  }{2017}]{pix2pix2016}
Isola, P.; Zhu, J.-Y.; Zhou, T.; and Efros, A.~A.
\newblock 2017.
\newblock Image-to-image translation with conditional adversarial networks.
\newblock In {\em CVPR}.

\bibitem[\protect\citeauthoryear{Jin, Lu, and Ruan}{2015}]{jin2015coupled}
Jin, Y.; Lu, J.; and Ruan, Q.
\newblock 2015.
\newblock Coupled discriminative feature learning for heterogeneous face
  recognition.
\newblock {\em TIFS} 10(3):640--652.

\bibitem[\protect\citeauthoryear{Juefei-Xu, Pal, and
  Savvides}{2015}]{juefei2015cvprw_nir}
Juefei-Xu, F.; Pal, D.~K.; and Savvides, M.
\newblock 2015.
\newblock Nir-vis heterogeneous face recognition via cross-spectral joint
  dictionary learning and reconstruction.
\newblock In {\em CVPRW},  141--150.

\bibitem[\protect\citeauthoryear{Kan, Shan, and Chen}{2016}]{kan2016cross-view}
Kan, M.; Shan, S.; and Chen, X.
\newblock 2016.
\newblock Multi-view deep network for cross-view classification.
\newblock In {\em CVPR},  4847--4855.

\bibitem[\protect\citeauthoryear{Klare and Jain}{2010}]{klare2010heterogeneous}
Klare, B., and Jain, A.~K.
\newblock 2010.
\newblock Heterogeneous face recognition: Matching nir to visible light images.
\newblock In {\em ICPR},  1513--1516.

\bibitem[\protect\citeauthoryear{Ledig \bgroup et al\mbox.\egroup
  }{2017}]{ledig2016photo}
Ledig, C.; Theis, L.; Husz{\'a}r, F.; Caballero, J.; Cunningham, A.; Acosta,
  A.; Aitken, A.; Tejani, A.; Totz, J.; Wang, Z.; et~al.
\newblock 2017.
\newblock Photo-realistic single image super-resolution using a generative
  adversarial network.
\newblock In {\em CVPR}.

\bibitem[\protect\citeauthoryear{Lei and Li}{2009}]{lei2009coupled}
Lei, Z., and Li, S.~Z.
\newblock 2009.
\newblock Coupled spectral regression for matching heterogeneous faces.
\newblock In {\em CVPR},  1123--1128.

\bibitem[\protect\citeauthoryear{Lei \bgroup et al\mbox.\egroup
  }{2008}]{lei2008CCA_mapping}
Lei, Z.; Bai, Q.; He, R.; and Li, S.~Z.
\newblock 2008.
\newblock Face shape recovery from a single image using cca mapping between
  tensor spaces.
\newblock In {\em CVPR},  1--7.

\bibitem[\protect\citeauthoryear{Lei \bgroup et al\mbox.\egroup
  }{2012}]{lei2012coupled}
Lei, Z.; Liao, S.; Jain, A.~K.; and Li, S.~Z.
\newblock 2012.
\newblock Coupled discriminant analysis for heterogeneous face recognition.
\newblock {\em TIFS} 7(6):1707--1716.

\bibitem[\protect\citeauthoryear{Lezama, Qiu, and Sapiro}{2016}]{lezama2016not}
Lezama, J.; Qiu, Q.; and Sapiro, G.
\newblock 2016.
\newblock Not afraid of the dark: Nir-vis face recognition via cross-spectral
  hallucination and low-rank embedding.
\newblock In {\em CVPR}.

\bibitem[\protect\citeauthoryear{Li \bgroup et al\mbox.\egroup
  }{2013}]{li2013casia}
Li, S.; Yi, D.; Lei, Z.; and Liao, S.
\newblock 2013.
\newblock The casia nir-vis 2.0 face database.
\newblock In {\em CVPRW},  348--353.

\bibitem[\protect\citeauthoryear{Li \bgroup et al\mbox.\egroup
  }{2017}]{li2017perceptual}
Li, J.; Liang, X.; Wei, Y.; Xu, T.; Feng, J.; and Yan, S.
\newblock 2017.
\newblock Perceptual generative adversarial networks for small object
  detection.
\newblock In {\em CVPR}.

\bibitem[\protect\citeauthoryear{Liao \bgroup et al\mbox.\egroup
  }{2009}]{liao2009heterogeneous}
Liao, S.; Yi, D.; Lei, Z.; Qin, R.; and Li, S.~Z.
\newblock 2009.
\newblock Heterogeneous face recognition from local structures of normalized
  appearance.
\newblock In {\em ICB},  209--218.

\bibitem[\protect\citeauthoryear{Lin and Tang}{2006}]{lin_tang2006}
Lin, D., and Tang, X.
\newblock 2006.
\newblock Inter-modality face recognition.
\newblock In {\em ECCV},  13--26.

\bibitem[\protect\citeauthoryear{Liu \bgroup et al\mbox.\egroup
  }{2005}]{liu2005nonlinear}
Liu, Q.; Tang, X.; Jin, H.; Lu, H.; and Ma, S.
\newblock 2005.
\newblock A nonlinear approach for face sketch synthesis and recognition.
\newblock In {\em CVPR}, volume~1,  1005--1010.

\bibitem[\protect\citeauthoryear{Liu \bgroup et al\mbox.\egroup
  }{2016}]{liu2016transferring}
Liu, X.; Song, L.; Wu, X.; and Tan, T.
\newblock 2016.
\newblock Transferring deep representation for nir-vis heterogeneous face
  recognition.
\newblock In {\em ICB},  1--8.

\bibitem[\protect\citeauthoryear{Long \bgroup et al\mbox.\egroup
  }{2016}]{long2016RTN}
Long, M.; Zhu, H.; Wang, J.; and Jordan, M.~I.
\newblock 2016.
\newblock Unsupervised domain adaptation with residual transfer networks.
\newblock In {\em NIPS},  136--144.

\bibitem[\protect\citeauthoryear{Shao and Fu}{2017}]{shao2017cross}
Shao, M., and Fu, Y.
\newblock 2017.
\newblock Cross-modality feature learning through generic hierarchical
  hyperlingual-words.
\newblock {\em TNNLS} 28(2):451--463.

\bibitem[\protect\citeauthoryear{Shrivastava \bgroup et al\mbox.\egroup
  }{2017}]{shrivastava2016learning}
Shrivastava, A.; Pfister, T.; Tuzel, O.; Susskind, J.; Wang, W.; and Webb, R.
\newblock 2017.
\newblock Learning from simulated and unsupervised images through adversarial
  training.
\newblock In {\em CVPR}.

\bibitem[\protect\citeauthoryear{Socolinsky and
  Selinger}{2002}]{socolinsky2002TIR_analysis}
Socolinsky, D.~A., and Selinger, A.
\newblock 2002.
\newblock A comparative analysis of face recognition performance with visible
  and thermal infrared imagery.
\newblock Technical report, DTIC Document.

\bibitem[\protect\citeauthoryear{Tang and Wang}{2004}]{tang2004face_sketch}
Tang, X., and Wang, X.
\newblock 2004.
\newblock Face sketch recognition.
\newblock {\em TCSVT} 14(1):50--57.

\bibitem[\protect\citeauthoryear{Wang and
  Tang}{2009}]{wang_tang2009photo-sketch}
Wang, X., and Tang, X.
\newblock 2009.
\newblock Face photo-sketch synthesis and recognition.
\newblock {\em TPAMI} 31(11):1955--1967.

\bibitem[\protect\citeauthoryear{Wang \bgroup et al\mbox.\egroup
  }{2009}]{wang2009analysis}
Wang, R.; Yang, J.; Yi, D.; and Li, S.~Z.
\newblock 2009.
\newblock An analysis-by-synthesis method for heterogeneous face biometrics.
\newblock In {\em ICB},  319--326.

\bibitem[\protect\citeauthoryear{Wang \bgroup et al\mbox.\egroup
  }{2012}]{wang2012sketch_synthesis}
Wang, S.; Zhang, L.; Liang, Y.; and Pan, Q.
\newblock 2012.
\newblock Semi-coupled dictionary learning with applications to image
  super-resolution and photo-sketch synthesis.
\newblock In {\em CVPR},  2216--2223.

\bibitem[\protect\citeauthoryear{Wang \bgroup et al\mbox.\egroup
  }{2013}]{wangkaiye2013iccv}
Wang, K.; He, R.; Wang, W.; Wang, L.; and Tan, T.
\newblock 2013.
\newblock Learning coupled feature spaces for cross-modal matching.
\newblock In {\em ICCV},  2088--2095.

\bibitem[\protect\citeauthoryear{Wang, Shrivastava, and
  Gupta}{2017}]{wang2017fast}
Wang, X.; Shrivastava, A.; and Gupta, A.
\newblock 2017.
\newblock A-fast-rcnn: Hard positive generation via adversary for object
  detection.
\newblock In {\em CVPR}.

\bibitem[\protect\citeauthoryear{Wu, He, and Sun}{2015}]{wu2015lightened}
Wu, X.; He, R.; and Sun, Z.
\newblock 2015.
\newblock A lightened cnn for deep face representation.
\newblock {\em arXiv preprint arXiv:1511.02683}.

\bibitem[\protect\citeauthoryear{Yi \bgroup et al\mbox.\egroup
  }{2007}]{yi2007face_NIS-VIR}
Yi, D.; Liu, R.; Chu, R.; Lei, Z.; and Li, S.~Z.
\newblock 2007.
\newblock Face matching between near infrared and visible light images.
\newblock In {\em ICB},  523--530.

\bibitem[\protect\citeauthoryear{Yi \bgroup et al\mbox.\egroup
  }{2009}]{yi2009partial_NIS-VIR}
Yi, D.; Liao, S.; Lei, Z.; Sang, J.; and Li, S.~Z.
\newblock 2009.
\newblock Partial face matching between near infrared and visual images in mbgc
  portal challenge.
\newblock In {\em ICB},  733--742.

\bibitem[\protect\citeauthoryear{Yi, Lei, and Li}{2015}]{yidong2015shared}
Yi, D.; Lei, Z.; and Li, S.~Z.
\newblock 2015.
\newblock Shared representation learning for heterogenous face recognition.
\newblock In {\em FG}, volume~1,  1--7.

\bibitem[\protect\citeauthoryear{Zellinger \bgroup et al\mbox.\egroup
  }{2017}]{zellinger2017CMD}
Zellinger, W.; Grubinger, T.; Lughofer, E.; Natschl{\"a}ger, T.; and
  Saminger-Platz, S.
\newblock 2017.
\newblock Central moment discrepancy (cmd) for domain-invariant representation
  learning.
\newblock In {\em ICLR}.

\bibitem[\protect\citeauthoryear{Zhang, Wang, and
  Tang}{2011}]{zhang2011coupled}
Zhang, W.; Wang, X.; and Tang, X.
\newblock 2011.
\newblock Coupled information-theoretic encoding for face photo-sketch
  recognition.
\newblock In {\em CVPR},  513--520.

\bibitem[\protect\citeauthoryear{Zhu \bgroup et al\mbox.\egroup
  }{2014}]{zhu2014matching}
Zhu, J.-Y.; Zheng, W.-S.; Lai, J.-H.; and Li, S.~Z.
\newblock 2014.
\newblock Matching nir face to vis face using transduction.
\newblock {\em TIFS} 9(3):501--514.

\bibitem[\protect\citeauthoryear{Zhu \bgroup et al\mbox.\egroup
  }{2017}]{zhu2017unpaired}
Zhu, J.-Y.; Park, T.; Isola, P.; and Efros, A.~A.
\newblock 2017.
\newblock Unpaired image-to-image translation using cycle-consistent
  adversarial networks.
\newblock In {\em ICCV}.

\end{thebibliography}
\bibliographystyle{aaai}
\end{document}